%% file: plantclef2021.tex
\documentclass[
]{ceurart}

\usepackage{minted}
\setminted{breaklines=true}

\begin{document}

\copyrightyear{2021}
\copyrightclause{Copyright for this paper by its authors.
  Use permitted under Creative Commons License Attribution 4.0
  International (CC BY 4.0).}

\conference{CLEF 2021 -- Conference and Labs of the Evaluation Forum, September 21--24, 2021, Bucharest, Romania}

\title{Overview of PlantCLEF 2021: cross-domain plant identification}

\author[1]{Herv\'e Go\"eau}[%
orcid=0000-0003-3296-3795,
email=herve.goeau@cirad.fr
]

\author[1]{Pierre Bonnet}[%
orcid=0000-0002-2828-4389,
email=pierre.bonnet@cirad.fr
]

\author[2]{Alexis Joly}[%
orcid=0000-0002-2161-9940,
email=alexis.joly@inria.fr
]
\address[1]{CIRAD, UMR AMAP, Montpellier, Occitanie, France}
\address[2]{Inria, LIRMM, Univ Montpellier, CNRS, Montpellier, France}

\begin{abstract}
Automated plant identification has improved considerably thanks to recent advances in deep learning and the availability of training data with more and more field photos. However, this profusion of data concerns only a few tens of thousands of species, mainly located in North America and Western Europe, much less in the richest regions in terms of biodiversity such as tropical countries. On the other hand, for several centuries, botanists have systematically collected, catalogued and stored plant specimens in herbaria, especially in tropical regions, and recent efforts by the biodiversity informatics community have made it possible to put millions of digitised records online. The LifeCLEF 2021 plant identification challenge (or "PlantCLEF 2021") was designed to assess the extent to which automated identification of flora in data-poor regions can be improved by using herbarium collections. It is based on a dataset of about 1,000 species mainly focused on the Guiana Shield of South America, a region known to have one of the highest plant diversities in the world. The challenge was evaluated as a cross-domain classification task where the training set consisted of several hundred thousand herbarium sheets and a few thousand photos to allow learning a correspondence between the two domains. In addition to the usual metadata (location, date, author, taxonomy), the training data also includes the values of 5 morphological and functional traits for each species. The test set consisted exclusively of photos taken in the field. This article presents the resources and evaluations of the assessment carried out, summarises the approaches and systems used by the participating research groups and provides an analysis of the main results.
\end{abstract}


\maketitle

\begin{figure}[!t]
    \centering
    \begin{tabular}{cc}
         \includegraphics[height=0.4\linewidth]{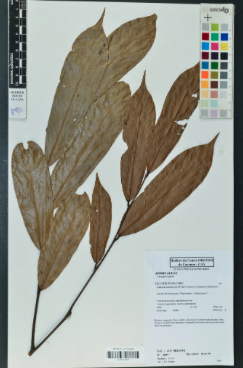}
        & \includegraphics[width=0.4\linewidth]{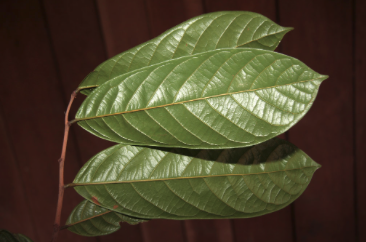} 
    \end{tabular}
\caption{An herbarium sheet and a field photo of the same individual plant (\textit{Unonopsis stipitata Diels}). Despite the very different visual appearances between the two types of images such as light reflection and leaf color, the overall leaf shape, vein structure, and leaf insertion on the branch remain invariant in both domains.}
\label{fig:illustrationspecimensphotoherbarium}
\end{figure}

Automated identification of the living world has improved considerably in recent years. In particular, in the LifeCLEF 2017 Plant Identification challenge, impressive identification performances have been measured with recent deep learning models (e.g. up to 90\% classification accuracy on 10,000 species), and it was shown in \cite{lifeclef2018} that automated systems are today not so far from human expertise. However, these conclusions are only valid for species that live predominantly in Europe and North America. Therefore, the LifeCLEF 2019 plant identification challenge focused on tropical countries, where there are generally far fewer observations and images collected and where the flora is much more difficult for human experts to identify.\\
In the meantime, biodiversity informatics initiatives such as iDigBio\footnote{\label{note1}\url{http://portal.idigbio.org/portal/search}} or e-ReColNat\footnote{\label{note2}\url{https://explore.recolnat.org/search/botanique/type=index}} have made millions of digitized herbarium sheets stored in many natural history museums over the world, available online. Over more than 3 centuries, generations of botanists have systematically collected, catalogued and stored plant specimens in herbaria. These specimens have great scientific value and are regularly used to study species variability, phylogenetic relationships, evolution or phenological trends. In particular, one of the key step in the work of botanists and taxonomists is to find the herbarium sheets that correspond to a new specimen observed in the field. This task requires a high level of expertise and can be very tedious. The development of automated tools to facilitate this work is therefore of crucial importance.\\ 
Following on from the PlantCLEF challenges held in previous years \cite{plantclef2011,plantclef2012,plantclef2013,plantclef2014,plantclef2015,plantclef2016,plantclef2017,expertlifeclef2018,plantclef2019}, a new challenge was introduced in 2020 and designed to assess the extent to which automated identification on flora in data-deficient regions can be improved by using natural history collections of herbarium sheets. In tropical countries, many species are not easily available, resulting in a very limited number of field-collected photographs, whereas several hundred or even several thousand herbarium sheets have been collected over the centuries. Herbarium collections potentially represent a large amount of data to train species prediction models, but they also induce a much more difficult problem, usually called \textit{cross-domain classification task}. Indeed, a plant photographed in the field may have a very different visual appearance than its dried version placed on a herbarium sheet (as illustrated in Figure \ref{fig:illustrationspecimensphotoherbarium}).\\ 

\section{Datasets and task description}
\label{Dataset}

\subsection{Training set}

\subsubsection{Visual content}

\begin{figure}
\centering
\includegraphics[width=\linewidth]{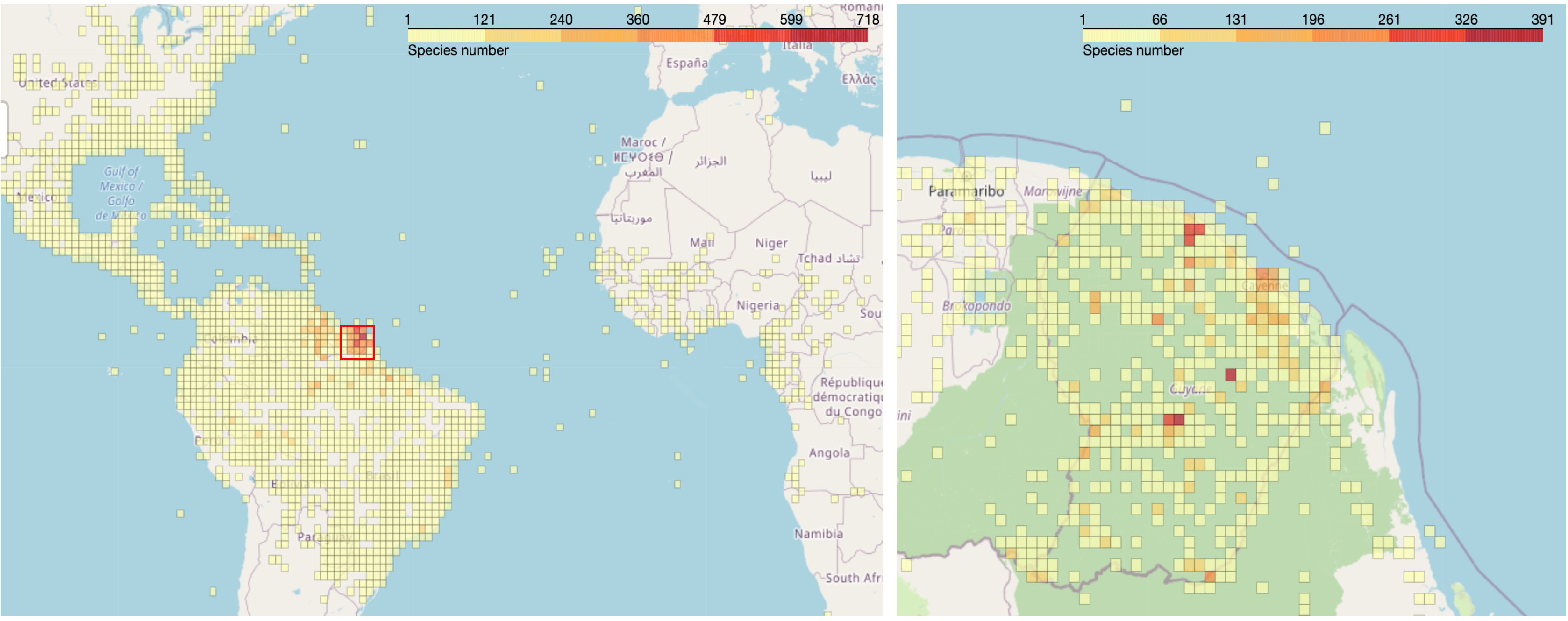}
\caption{Density grid maps by number of species of geolocated plant specimens in the PlantCLEF2021 dataset. Many species have also been collected in other regions outside French Guiana, over a large part of the Americas, but also in Africa for some of them.}
\label{fig:plantclef2020speciesmap}
\end{figure}

The task was based on a dataset of 997 species mainly focused on the Guiana shield and the Northern Amazon rainforest (see Figure \ref{fig:plantclef2020speciesmap}), an area known to have one of the greatest diversity of plants and animals in the world. The dataset contains 321,270 herbarium sheets (see Table\ref{tab:datasetorigin} for detailed information). About 12\% were collected in French Guyana and hosted in the "Herbier IRD de Guyane" (IRD Herbarium of French Guyana). These herbarium sheets were digitized in the context of the e-ReColNat\textsuperscript{\ref{note2}} project. The remaining herbarium sheets come from the iDigBio\textsuperscript{\ref{note1}} portal (the US National Resource for Advancing Digitization of Biodiversity Collections). 

In order to enable learning a mapping between the two domains (i.e. between the "source" domain of herbarium sheets and the "target" domain of field photos), a relatively smaller set of 6,316 photos in the field was provided additionally to the large herbarium sheets dataset. About 62 \% of them also come from he iDigBio portal and were acquired by various photographers related to numerous institutes and national museums that share their data in iDigBio. Besides, two highly trusted experts of the French Guyana flora, Marie-Françoise Pr\'evost "Fanchon" \cite{delprete2013marie} and Jean-François Molino\footnote{\url{https://scholar.google.fr/citations?user=xZXYc4kAAAAJ&hl=fr}} provided the remaining field photos that were divided between the training set and the test set.

A valuable asset of the training set is that a set of 354 plant observations are provided with both herbarium sheets and field photos for the same individual plant. This potentially allows a more precise mapping between the two domains (see previous Figure \ref{fig:illustrationspecimensphotoherbarium} as an example).\\
It should also be noted that about half of the species in the training set (495 to be precise) is only represented by herbarium sheets which makes training a model even more difficult without the presence of field photo examples.

\begin{table}[!t]
    \caption{Details of the PlantCLEF 2021 dataset according to the origin of the pictures and their domain.}
    \centering
    \begin{tabular}{cccrr}
         Origin                 & Domain            & Used as   &   \#Pictures  &   \#Species\\
         \hline
         Herbier IRD de Guyane  & Herbarium sheets  & Train     &   38,552  &   631\\
         iDigBio                & Herbarium sheets  & Train     &   282,718 &   991\\
         iDigBio                & Field photos      & Train     &   3,935   &   426\\
         Fanchon                & Field photos      & Train     &   1,130   &   183\\
         Molino                 & Field photos      & Train     &   1,251   &   125\\
         \hline
         Fanchon                & Field photos      & Test      &   1,830   &   271\\
         Molino                 & Field photos      & Test      &   1,356   &   166\\
         \hline
         \hline
         Train (all) & Herbarium sheets & Train & 321,270 & 997\\
         Train (all) & Field photos & Train & 6,316 & 502\\
         Test (all) & Field photos & Test & 3,186 & 408\\
         \hline
    \end{tabular}
\label{tab:datasetorigin}
\end{table}

\subsection{Traits metadata}

Additional metadata at the species level expressing functional traits were introduced this year. This is a very valuable information that can potentially help improve prediction models. Indeed, it can be assumed that species which share the same functional traits also share to some extent common visual appearances. This information can then potentially be used to guide the training of a model through auxiliary loss functions for instance. The information was collected through the Encyclopedia of Life API. The 5 most comprehensive traits have been verified and completed by experts in Guyanese flora, so that each species has a value for each trait. Below we list the names of the 5 traits as well as the possible values associated with these traits.\\
\\
\textbf{Plant growth form:} describes which plant growth form can take a species among these 4 possibilities: climber, herb, shrub, tree. It is important to note that a species can sometimes be associated with several forms of growth. For example, a young plant of the species \textit{Justicia betonica L.} can be considered as an herb while in adulthood it would be described as a shrub.\\
\\
\textbf{Habitat:} a set non standardised free tag(s) describing the typical habitats of a given species. As examples, we can indicate the most frequently used tags: tropical, moist, broadleaf, forest, flooded, grassland, rocky, non-wetland, savanna, shrubland, coastal.\\
\\
\textbf{Plant lifeform:} refers to the physical support of development used by a species. Below is a list of all possible values and their definitions:
\begin{itemize}
    \item Aquatic plant.
    \item Epiphyte: an organism that grows on the surface of a plant and derives its moisture and nutrients from the air, rain, water (in marine environments) or from debris accumulating around it.
    \item Geophyte: species that develop organs for storing energy (water or carbohydrates).
    \item Helophyte: a plant that grows in or near water and is either emergent, submergent, or floating.
    \item Hemiepiphyte: a plant that spends part of its life cycle as an epiphyte.
    \item Hydrophyte: close to helophyte.
    \item Lithophyte: plants that grow in or on rocks.
    \item Pleustophyte: a plant living in the thin surface layer existing at the air-water interface of a body of water which serves as their habitat.
    \item Succulent plant: a plant with parts that are thickened, fleshy, and engorged, usually to retain water in arid climates or soil conditions (close to geophyte).
    \item Terrestrial plant.
\end{itemize}

\noindent \textbf{Trophic guild:} can be common to any group of species that exploit the same resources, or that exploit different resources in related ways. 
\begin{itemize}
    \item Carnivorous plant.
    \item Parasite: a plant that derives some or all of its nutritional requirement from another living plant.
    \item Hemiparasite: partially parasite.
    \item Photoautotroph: a plant that is capable of synthesizing its own food from inorganic substances using light as an energy source
    \item Saprotrophic: a plant which secrete digestive juices in dead and decaying matter and convert it into a solution and absorb it.
\end{itemize}

\noindent \textbf{Woodiness:} expresses whether the species is capable of producing "lignin" (wood). Then, the values are basically herb or woody.

\subsection{Test set}
The test set was composed of 3,186 photos in the field related to 638 plant observations (about 5 pictures per plants on average). To avoid bias related to similar pictures coming from neighboring plants in the same observation site, we ensured that all observations of a given species by a given collector were either in the training set or in the test set but never spread over the two sets. For instance, for the observations of J.F. Molino, the 166 species in the test set are different from the 125 species in the training set.

\begin{figure}[!t]
\centering
\includegraphics[width=0.92\linewidth]{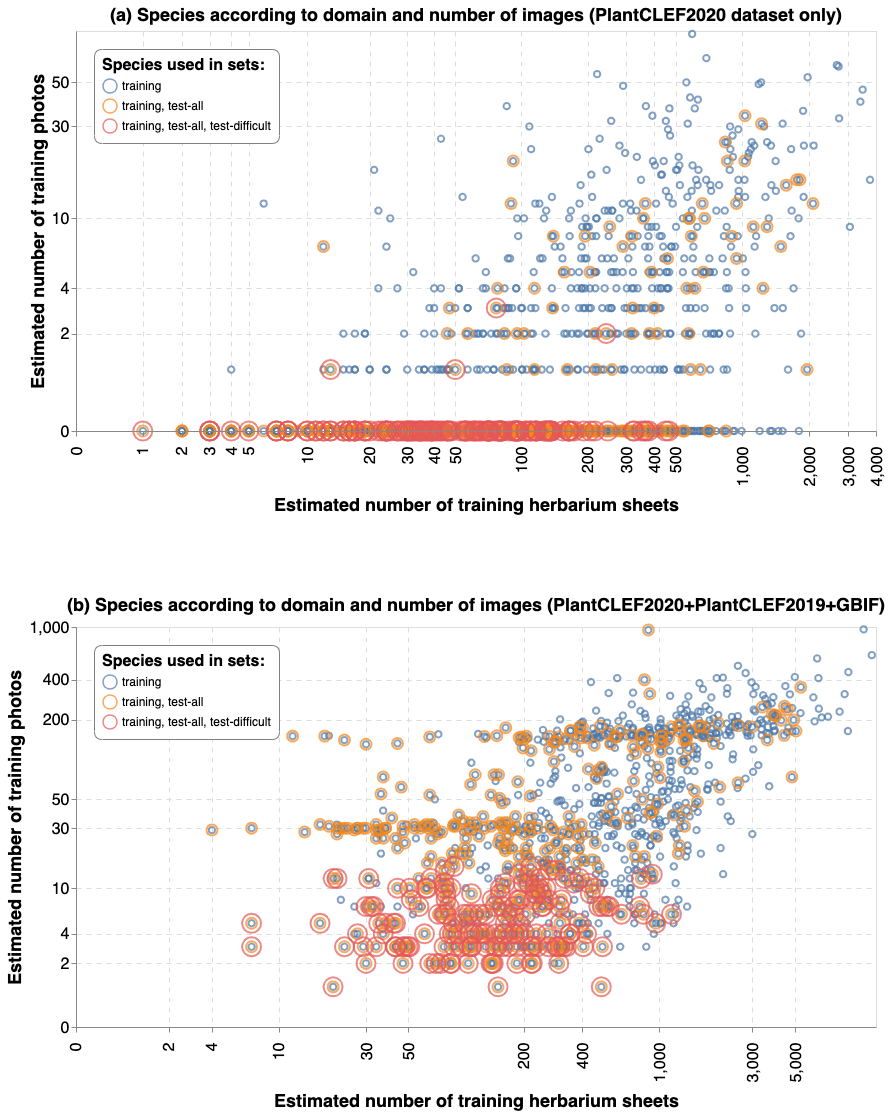}
\caption{Species according to the estimated number of images for each domain in the training set (in blue). Each species is surrounded by an additional orange circle if it is used in the test set, and a red circle if used in the test subset of difficult species (with few field photos according to the PlantCLEF 2021 the training set). The bottom graph revises the positions of the species by including additional training pictures from external datasets that could be used by the participants. It is estimated that most of the species related to the difficult test subset have less than 10 field photos.}
\label{fig:plantclef2020speciesdistrib}
\end{figure}

Most importantly, plant species in the test set were selected according to the number of field photos illustrating them in the training set. As it can be observed in Figure \ref{fig:plantclef2020speciesdistrib} (a), the priority was given to species with few or no field pictures at all. Such a choice may seem drastic, making the task extremely difficult, but the underlying idea was to encourage and promote methods that are as generic as possible, capable of transferring knowledge between the two domains, even without any examples in the target domain for some classes. The second motivation of this choice, was to impose a mapping between herbarium and field photos and avoid that classical methods based on CNNs perform well because of an abundance of field photos in the training set rather than the use of herbarium sheets.

\subsection{External training sets}

Participants to the evaluation were allowed to use complementary training data (e.g. for pre-training purposes) but on the condition that (i) the experiment is entirely reproducible, i.e. that the used external resource is clearly referenced and accessible to any other research group, (ii) the use of external training data or not is clearly mentioned for each evaluated method, and (iii) the additional resource does not contain any of the test observations. External training data was thus allowed but participants had to provide at least one submission that used only the training data provided this year. The organizers suggested two external datasets used in the previous edition PlantCLEF2019 \cite{plantclef2019}, \cite{cmp2019lifeclef}.

\subsection{Task Description}
\noindent \textbf{Evaluation metrics:} the goal of the task was to identify the correct species of the 638 plant of the test set. For every plant, the evaluated systems had to return a list of species, ranked without ex-aequo. Each participating group was allowed to submit up to 10 \textit{run files} built from different methods or systems (a \textit{run file} is a formatted text file containing the species predictions for all test items).

The main evaluation measure for the challenge was the Mean Reciprocal Rank (MRR), which is defined as the mean of the multiplicative inverse of the rank of the correct answer:
$$ \frac{1}{Q} \sum_{q=1}^Q \frac{1}{\text{rank}_q} $$ 
where $Q$ is the number of plant observations and $\mathrm{rank}_q$ is the predicted rank of the true label for the $q$th plant observation.

A second evaluation measure was again the MRR but computed on a subset of observations of difficult species that are rarely photographed in the field. Species were selected based on the most comprehensive estimate of the number of field photos from different data sources (iDigBio, GBIF, Encyclopedia of Life, Bing and Google Image search engines). It is therefore a more challenging metric because it focuses on the species which impose a mapping between herbarium and field photos. Figure \ref{fig:plantclef2020speciesdistrib} (b) revises the previous Figure \ref{fig:plantclef2020speciesdistrib} (a) according to the considered external data sources and shows that many plant observations in the difficult test subset are related to species estimated to have less than 10 field photos.\\
\\
\textbf{Course of the challenge}: the training data was publicly shared mid February 2021 through the AICrowd platform\footnote{\url{https://www.aicrowd.com/challenges/lifeclef-2021-plant}}. Any research team wishing to participate in the evaluation could register on the platform and download the data. The test data was shared in mid-April but without the species labels, which were kept secret. Each team could then submit up to 10 submissions corresponding to different methods or different settings of the same method. A submission (also called a \textit{run}) takes the form of a csv file containing the predictions of the method being evaluated for all observations in the test set. For each submission, the calculation of the evaluation metrics is then done automatically and visible to the participant. Once, the submission phase was closed (mid May), the participants could also see the evaluation metric values of the other participants. As a last important step, each participant was asked to provide a \textit{working note}, i.e. a detailed technical report containing all technical information required to reproduce the results of all submissions. All LifeCLEF \textit{working notes} are reviewed by at least two members of LifeCLEF organizing committee to ensure a sufficient level of quality and reproducibility.

\section{Participants and methods}
\label{participants}
About 30 teams/researchers registered for the PlantCLEF challenge 2021 and 4 of them finally submitted runs. Details of the methods are developed in the individual working notes of the participants (NeuonAI \cite{NeuonAI2021}, Lehigh University \cite{LehighPlantCLEF2021}). The other teams did not provide a detailed description of their systems, but some informal descriptions were sometimes provided in the metadata associated with the submissions and partially contributed to the comments below.\\
\\
\textbf{Neuon AI, Malaysia, 7 runs \cite{NeuonAI2021}}: the method is actually an extension of a previous approach already successfully evaluated last year in PlantCLEF2020 \cite{NeuonAI2020}. It is based on a two-streamed Herbarium-Field Triplet Loss Network (HTFL) to assess the similarity between herbarium and field pairs thus matching species from both herbarium and field domains. This mechanism is better at predicting observations of species with missing field images in the training set than traditional CNNs. The general concept is to train the network with triplet samples (2 samples from the same species and 1 from another species) to minimize the distance between the same species and maximize the distance between different species. Then, an herbarium dictionary is computed: each class (each specie)s is associated with an unique embedding computed as the average of the random herbarium sheets from the class. For inference, a plant observation is then associated with a unique embedding computed as the average of the embeddings of all (augmented) field photos of the observation. Cosine similarity is used as a distance metric between the embeddings of all herbarium classes and the embedding of the tested field observation. It is then transformed with inverse distance weighting into probabilities to rank the classes.\\
The authors introduced this year several novelties to improve the method: 
\begin{itemize}
    \item They trained a complementary One-Streamed Mixed network (OSM) by taking both herbarium and field images as input to learn the features of each species irrespective of their domains. The learned features of the OSM network are used as a way to measure the feature similarity between herbarium-field pairs instead of directly classifying them. This mechanism allows for the prediction of classes where field images are missing in a similar way triplet network does. 
    \item Complementary dictionary for embedding: they showed that by adding field images to the herbarium images for computing the herbarium dictionary, the “HFTL (field)” performs better on difficult “unseen” species without field images in the training set, but worse for species having field images in the training than the initial HTFL approach using only herbarium sheets for computing the dictionary.
    \item Ensemble: instead of opposing the two ways of building dictionaries, they combine the two approaches through an ensemble of networks. Finally, the best performances are obtained when combining the HTLF, HTFL(field) and also OSM, all being built and declined with two CNN architectures (Inception-v4 and Inception-ResNet-v2).
\end{itemize}

\noindent \textbf{LU, Lehigh University, USA, 9 runs \cite{LehighPlantCLEF2021}}: these participants started from the fact that although many methods are proposed for domain adaptation, most of them are tested on small domain divergence datasets, which may have lower transferability to large-divergence datasets, and the data imbalance problem is not well addressed. To address these challenges, this participant proposes to extend the CORAL loss \cite{sun2016deep} designed to align the distribution of the two domains with two contributions:
\begin{itemize}
    \item A weighted cross-entropy loss (optimized on the labeled samples and pseudo-labeled samples) allowing to take into account the imbalance distribution of the classes
    \item A filtering of the used pseudo-labels to focus only on the ones with a sufficient degree of confidence. The confidence threshold decreases over time to integrate progressively more difficult test samples.
\end{itemize}
Unlike the approach developed by NeuonAI or the organizer's submissions (see below), these participants do not seem to have used external data, which may partially explain the difference in performance despite a promising and interesting approach.
Moreover, we can notice that both Neuon and LU teams did not exploit in their approaches the new trait metadata introduced this year.\\

\noindent \textbf{Organizer's submissions, 7 runs}: the purpose of these submissions was to measure whether or not the functional trait metadata introduced this year could help improve the performance of a system initially exploiting only image data. The submissions are based on the "winning" solution from the previous year designed by Juan Villacis, a former student from the TEC Costa Rica and the Pl@ntNet team \cite{ITCRPlantNet2020}. The method uses a Few Shot Adversarial Domain Adaptation approach \cite{fewshotda} (FSADA) where the purpose is to learn a domain agnostic feature space while preserving the discriminative ability of the features for performing the species classification task. First, a ResNet50 is finetuned in the herbarium sheets only and used then as an encoder to extract features on both herbarium sheets or field photos. Then, given random pairs of extracted features, a discriminator is trained to distinguish 4 categories: (1) different domains and different classes, (2) different domains and same class, (3) same domain and different classes, (4) same domain and same classes. Finally, during a last stage, the encoder, the discriminator and the classifier are trained together. Domain adaptation is achieved once the discriminator is not able to distinguish samples from categories (1) and (2) and categories (3) and (4), when the discriminator is not able to tell which was the original domain. The best single model was an extension with 3 classifiers and 3 discriminators using 3 taxonomic levels (species, genus, family), while using external datasets (PlantCLEF 2019\cite{plantclef2019} and GBIF \cite{cmp2019lifeclef}). This previous best solution was retrained and again submitted this year as "Organizer's submission 3" (submissions 1 \& 2 didn't use the genus and family levels, and submission 1 didn't use external data). Following the same idea, the organizer's submissions 4, 5, 6 extended the submission 3 by introducing additional discriminators and classifiers related to the traits, respectively "plant lifeform" (10 classes), "woodiness" (2 classes), "plant growthform" (using actually in this specific case 4 independent discriminators and binary classifiers related to the 4 values "climber", "herb", "shrub", "tree"). Submission 7 used a total of 10 discriminators and classifiers exploiting the 3 taxonomic levels and traits mentioned just before.

\section{Results}

\begin{figure}
\centering
\includegraphics[width=\linewidth]{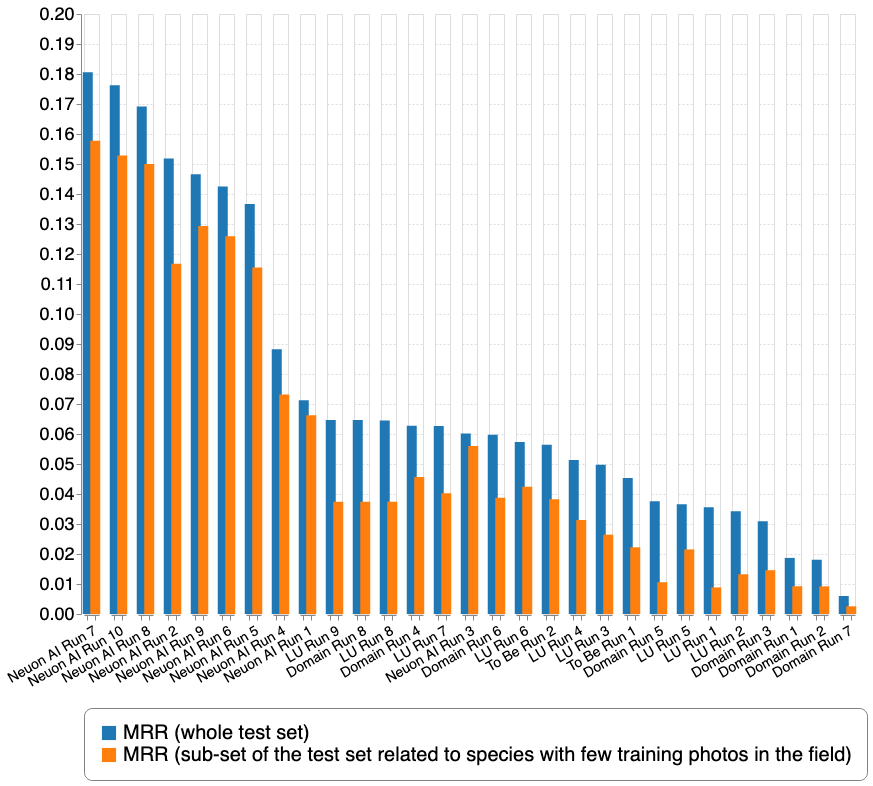}
\caption{PlantCLEF 2021 evaluation results sorted by the primary evaluation metric, i.e. the Mean Reciprocal Rank over the entire test set.}
\label{fig:PlantCLEF2021Scores}
\end{figure}

\begin{figure}
\centering
\includegraphics[width=\linewidth]{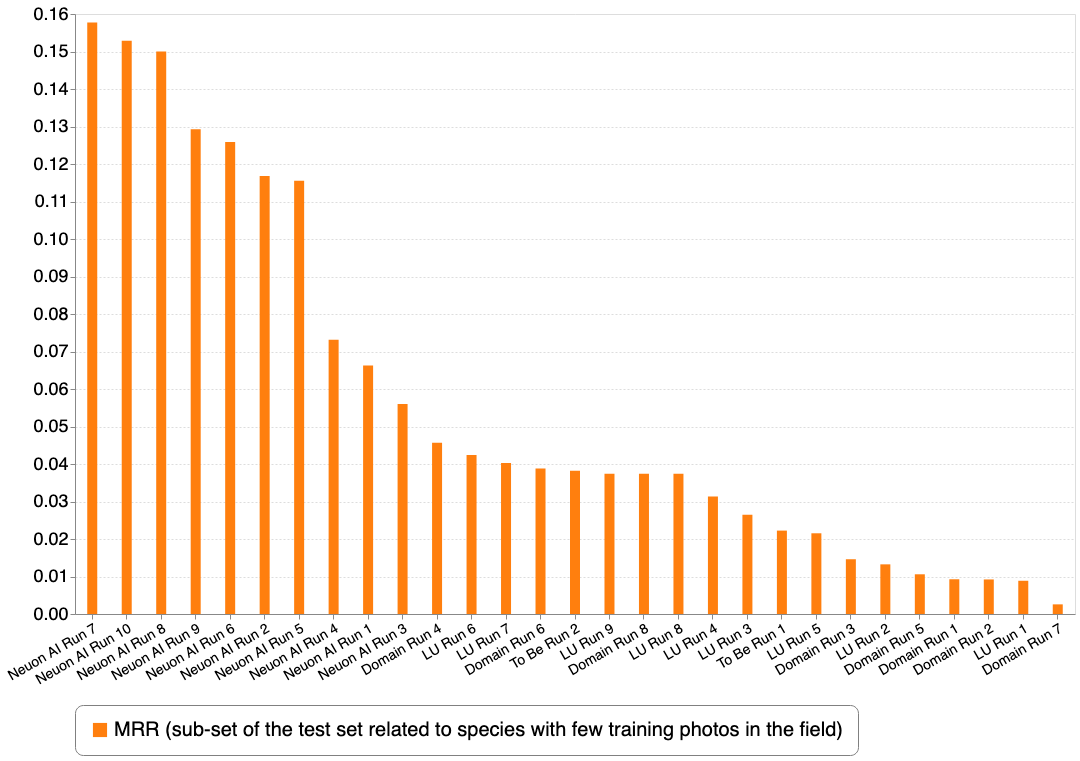}
\caption{PlantCLEF 2021 evaluation results based on the secondary metric, i.e. the Mean Reciprocal Rank over the subset of difficult species with few or no field photos in the training set.}
\label{fig:PlantCLEF2021ScoresSecondMetric}
\end{figure}

\begin{figure}
\centering
\includegraphics[width=\linewidth]{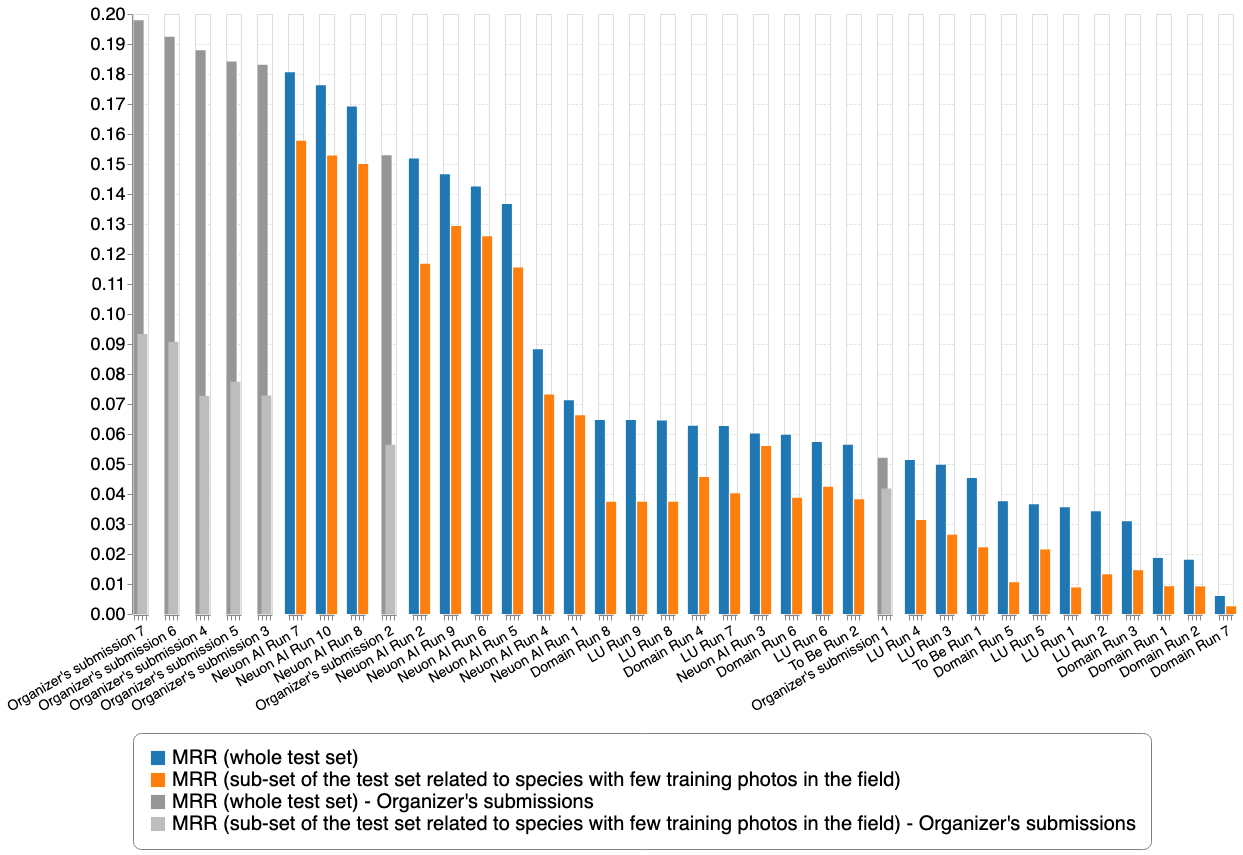}
\caption{PlantCLEF 2021 evaluation results sorted by the primary evaluation metric including the organizer's submissions.}
\label{fig:PlantCLEF2021AllWithOrganizersSubmissions}
\end{figure}

We report in Figure \ref{fig:PlantCLEF2021Scores} and Table \ref{tab:rawresults} the performances achieved by the 29 evaluated runs. Figure \ref{fig:PlantCLEF2021ScoresSecondMetric} reorganizes the results according to the second MRR metric focusing on the most difficult species, while Figure \ref{fig:PlantCLEF2021AllWithOrganizersSubmissions} adds the organizer's submissions in the initial Figure \ref{fig:PlantCLEF2021Scores}.\\
\input{tables/results_table.tex}


The main outcomes we can derive from that results are the following ones:\\

\noindent \textbf{The most difficult PlantCLEF challenge ever.} Traditional classification models based on CNNs perform very poorly on the task. Domain Adaptation methods (DA) based on CNNs perform much better but the task remains difficult even with these dedicated techniques. The best submitted run barely approaches a MRR of 0.2.\\

\noindent \textbf{External data improves DA approaches}. The best submissions from NeuonAI and the organizers used both complementary external data (the PlantCLEF2017 dataset \cite{plantclef2017} for NeuonAI while the organizer's submissions used a GBIF collection \cite{cmp2019lifeclef} and the PlantCLEF2019 dataset \cite{plantclef2019}). The impact of external data can be highlighted when comparing runs 1 \& 2 from the organizers for instance where the MRR increases from 0.052 to 0.153. LU team used only used the relatively small number of field photos provided in the PlantCLEF2021 dataset, which may partially explain why their submissions have lower performance.\\

\noindent \textbf{Genericity and stability.} Regarding the difference between the two MRR metrics (whole test set vs. difficult species), the NeuonAI team demonstrated that it is possible to achieve equivalent and quite good performance for all species, even those that have few or no field photos at all in the training dataset. This impressive genericity is mainly obtained by the use of an ensemble of several two-streamed Herbarium-Field Triplet Loss (HFTL) networks and several One-Streamed Mixed (OSM) networks. Rather than focusing on learning a common feature invariant domain as for the other team's submissions, the NeuonAI's approach focuses on a deep metric learning on features embeddings. 
For the ensemble, they combined through their submissions various combinations of (a) different CNN architectures (Inception-ResNet-v2 and Inception-V4), (b) whether or not to add to the herbarium pictures the field photos for the calculation of class embbedings, and (c) different levels of data augmentation. NeuonAI Run 7, which used 3 HTFL networks, all adding field photos for the class embbedings, and 2 OSM networks, achieved the best scores over the participants, for both MRR metrics. 
Looking solely at the the second MRR score, this approach seems to be more effective in transferring knowledge to the least frequently photographed species in the field, which was the most difficult goal to achieve.\\

\noindent \textbf{Multi-task approaches have a positive impact on performance}, especially when using the taxonomy \textbf{genus, family} and some of the species \textbf{traits}. By adding auxiliary tasks to the FSADA approach as mentioned in \cite{ITCRPlantNet2020}, the genus and family information help the discriminators and the classifiers to learn a better domain invariant feature space, and contribute to improve significantly the performance (see Organizer's submissions 2 \& 3 where the MRR increases from 0.153 to 0.183). Organizer's submissions 4, 5 and 6 extended this approach by adding an auxiliary task related to the trait species information. This appeared to also contribute to improve performance. All these auxiliary tasks based on traits contributed to improve the performances, even slightly as for the ``woodiness" trait.\\

\noindent \textbf{The most informative species trait is the ``plant growth form"}. Organizer's submissions 4, 5 and 6 demonstrate that adding auxiliary tasks based on species traits improves performance. As hypothesised, it seems to help gathering and discriminating wide groups of plant species sharing similar visual aspects (such as tendrils for climber plants, typical large leaves for tropical trees against smaller leaves for shrubs or long thin leaves and frequent flowers for herbs). 
Finally, the last organizer's submission 7 combining a total of 8 auxiliary tasks resulted in the highest score for the primary MRR metric over the challenge, but half as good as the best NeuonAI's submission regarding the second MRR metric focusing on the very difficult species to identify. 


\section{Conclusion}
This paper presented the overview and the results of the LifeCLEF 2021 plant identification challenge following the 10 previous editions conducted within CLEF evaluation forum. This year's task was particularly challenging, focusing on species rarely photographed in the field in the northern tropical Amazon. The results revealed that the last advances in domain adaptation enable the use of herbarium data to facilitate the identification of rare tropical species for which no or very few other training photos are available. A mapping domain adaptation technique based on a two-streamed Herbarium-Field triplet loss network reached an impressive genericity by obtaining quite high similar results regardless of whether the species have many or very few field photos in the training set. 
We believe that the proposed task may be in the future a new baseline dataset in the field of domain adaptation, and motivate new contributions through a realistic and crucial usage for the plant biology research community.

\begin{acknowledgments}
This project has received funding from the French National Research Agency under the Investments for the Future Program, referred as ANR-16-CONV-0004 and from the European Union’s Horizon 2020 research and innovation program under grant agreement No 863463 (Cos4Cloud project). This work was supported in part by the Microsoft AI for Earth program.
\end{acknowledgments}


\appendix


\end{document}

%% file: tables/results_table.tex
\begin{table}
\caption{Results of the LifeCLEF 2021 Plant Identification Task}
    \centering
    \begin{tabular}{ccc}
Team run    & MRR (whole test set)	&	MRR (difficult species)\\
\hline
Neuon AI Run 7 & 0.181 & 0.158 \\
Neuon AI Run 10 & 0.176 & 0.153 \\
Neuon AI Run 8 & 0.169 & 0.15 \\
Neuon AI Run 2 & 0.152 & 0.117 \\
Neuon AI Run 9 & 0.147 & 0.129 \\
Neuon AI Run 6 & 0.143 & 0.126 \\
Neuon AI Run 5 & 0.137 & 0.116 \\
Neuon AI Run 4 & 0.088 & 0.073 \\
Neuon AI Run 1 & 0.071 & 0.066 \\
LU Run 9  & 0.065 & 0.037 \\
Domain Run 8 & 0.065 & 0.037 \\
LU Run 8 & 0.065 & 0.037 \\
LU Run 7 & 0.063 & 0.04 \\
Domain Run 4 & 0.063 & 0.046 \\
Neuon AI Run 3 & 0.06 & 0.056 \\
Domain Run 6 & 0.06 & 0.039 \\
LU Run 6 & 0.057 & 0.042 \\
To Be Run 2 & 0.056 & 0.038 \\
LU Run 4 & 0.051 & 0.031 \\
LU Run 3 & 0.05 & 0.026 \\
To Be Run 1 & 0.045 & 0.022 \\
Domain Run 5 & 0.038 & 0.011 \\
LU Run 5 & 0.037 & 0.022 \\
LU Run 1 & 0.036 & 0.009 \\
LU Run 2 & 0.034 & 0.013 \\
Domain Run 3 & 0.031 & 0.015 \\
Domain Run 1 & 0.019 & 0.009 \\
Domain Run 2 & 0.018 & 0.009 \\
Domain Run 7 & 0.006 & 0.003 \\
\hline
Organizer's submission 7 & 0.198 & 0.093 \\
Organizer's submission 6 & 0.192 & 0.091 \\
Organizer's submission 4 & 0.188 & 0.073 \\
Organizer's submission 5 & 0.184 & 0.077 \\
Organizer's submission 3 & 0.183 & 0.073 \\
Organizer's submission 2 & 0.153 & 0.056 \\
Organizer's submission 1 & 0.052 & 0.042
\end{tabular}%
\label{tab:rawresults}
\end{table}

%% file: plantclef2021.bbl
\begin{thebibliography}{18}
\expandafter\ifx\csname natexlab\endcsname\relax\def\natexlab#1{#1}\fi
\providecommand{\url}[1]{\texttt{#1}}
\providecommand{\href}[2]{#2}
\providecommand{\path}[1]{#1}
\providecommand{\DOIprefix}{doi:}
\providecommand{\ArXivprefix}{arXiv:}
\providecommand{\URLprefix}{URL: }
\providecommand{\Pubmedprefix}{pmid:}
\providecommand{\doi}[1]{\href{http://dx.doi.org/#1}{\path{#1}}}
\providecommand{\Pubmed}[1]{\href{pmid:#1}{\path{#1}}}
\providecommand{\bibinfo}[2]{#2}
\ifx\xfnm\relax \def\xfnm[#1]{\unskip,\space#1}\fi
\bibitem[{Joly et~al.(2018)Joly, Go{\"e}au, Botella, Glotin, Bonnet, Vellinga, and M{\"u}ller}]{lifeclef2018}
\bibinfo{author}{A.~Joly}, \bibinfo{author}{H.~Go{\"e}au}, \bibinfo{author}{C.~Botella}, \bibinfo{author}{H.~Glotin}, \bibinfo{author}{P.~Bonnet}, \bibinfo{author}{W.-P. Vellinga}, \bibinfo{author}{H.~M{\"u}ller},
\newblock \bibinfo{title}{Overview of lifeclef 2018: a large-scale evaluation of species identification and recommendation algorithms in the era of ai},
\newblock in: \bibinfo{editor}{G.~J. Jones}, \bibinfo{editor}{S.~Lawless}, \bibinfo{editor}{J.~Gonzalo}, \bibinfo{editor}{L.~Kelly}, \bibinfo{editor}{L.~Goeuriot}, \bibinfo{editor}{T.~Mandl}, \bibinfo{editor}{L.~Cappellato}, \bibinfo{editor}{N.~Ferro} (Eds.), \bibinfo{booktitle}{{CLEF: Cross-Language Evaluation Forum for European Languages}}, volume \bibinfo{volume}{LNCS} of \textit{\bibinfo{series}{Experimental IR Meets Multilinguality, Multimodality, and Interaction}}, \bibinfo{publisher}{{Springer}}, \bibinfo{address}{Avigon, France}, \bibinfo{year}{2018}.
\bibitem[{Go{\"e}au et~al.(2011)Go{\"e}au, Bonnet, Joly, Boujemaa, Barth{\'e}l{\'e}my, Molino, Birnbaum, Mouysset, and Picard}]{plantclef2011}
\bibinfo{author}{H.~Go{\"e}au}, \bibinfo{author}{P.~Bonnet}, \bibinfo{author}{A.~Joly}, \bibinfo{author}{N.~Boujemaa}, \bibinfo{author}{D.~Barth{\'e}l{\'e}my}, \bibinfo{author}{J.-F. Molino}, \bibinfo{author}{P.~Birnbaum}, \bibinfo{author}{E.~Mouysset}, \bibinfo{author}{M.~Picard},
\newblock \bibinfo{title}{The imageclef 2011 plant images classification task},
\newblock in: \bibinfo{booktitle}{CLEF task overview 2011, CLEF: Conference and Labs of the Evaluation Forum, Sep. 2011, Amsterdam, Netherlands.}, \bibinfo{year}{2011}.
\bibitem[{Go{\"e}au et~al.(2012)Go{\"e}au, Bonnet, Joly, Yahiaoui, Barth{\'e}l{\'e}my, Boujemaa, and Molino}]{plantclef2012}
\bibinfo{author}{H.~Go{\"e}au}, \bibinfo{author}{P.~Bonnet}, \bibinfo{author}{A.~Joly}, \bibinfo{author}{I.~Yahiaoui}, \bibinfo{author}{D.~Barth{\'e}l{\'e}my}, \bibinfo{author}{N.~Boujemaa}, \bibinfo{author}{J.-F. Molino},
\newblock \bibinfo{title}{Imageclef2012 plant images identification task},
\newblock in: \bibinfo{booktitle}{CLEF task overview 2012, CLEF: Conference and Labs of the Evaluation Forum, Sep. 2012, Rome, Italy.}, \bibinfo{address}{Rome}, \bibinfo{year}{2012}.
\bibitem[{Go{\"e}au et~al.(2013)Go{\"e}au, Bonnet, Joly, Bakic, Barth{\'e}l{\'e}my, Boujemaa, and Molino}]{plantclef2013}
\bibinfo{author}{H.~Go{\"e}au}, \bibinfo{author}{P.~Bonnet}, \bibinfo{author}{A.~Joly}, \bibinfo{author}{V.~Bakic}, \bibinfo{author}{D.~Barth{\'e}l{\'e}my}, \bibinfo{author}{N.~Boujemaa}, \bibinfo{author}{J.-F. Molino},
\newblock \bibinfo{title}{The imageclef 2013 plant identification task},
\newblock in: \bibinfo{booktitle}{CLEF task overview 2013, CLEF: Conference and Labs of the Evaluation Forum, Sep. 2013, Valencia, Spain.}, \bibinfo{address}{Valencia}, \bibinfo{year}{2013}.
\bibitem[{Go{\"e}au et~al.(2014)Go{\"e}au, Joly, Bonnet, Selmi, Molino, Barth{\'e}l{\'e}my, and Boujemaa}]{plantclef2014}
\bibinfo{author}{H.~Go{\"e}au}, \bibinfo{author}{A.~Joly}, \bibinfo{author}{P.~Bonnet}, \bibinfo{author}{S.~Selmi}, \bibinfo{author}{J.-F. Molino}, \bibinfo{author}{D.~Barth{\'e}l{\'e}my}, \bibinfo{author}{N.~Boujemaa},
\newblock \bibinfo{title}{The lifeclef 2014 plant images identification task},
\newblock in: \bibinfo{booktitle}{CLEF task overview 2014, CLEF: Conference and Labs of the Evaluation Forum, Sep. 2014, Sheffield, United Kingdom.}, \bibinfo{address}{Sheffield, UK}, \bibinfo{year}{2014}.
\bibitem[{Go{\"e}au et~al.(2015)Go{\"e}au, Joly, and Bonnet}]{plantclef2015}
\bibinfo{author}{H.~Go{\"e}au}, \bibinfo{author}{A.~Joly}, \bibinfo{author}{P.~Bonnet},
\newblock \bibinfo{title}{Lifeclef plant identification task 2015},
\newblock in: \bibinfo{booktitle}{CLEF task overview 2015, CLEF: Conference and Labs of the Evaluation Forum, Sep. 2015, Toulouse, France.}, \bibinfo{year}{2015}.
\bibitem[{Go{\"e}au et~al.(2016)Go{\"e}au, Bonnet, and Joly}]{plantclef2016}
\bibinfo{author}{H.~Go{\"e}au}, \bibinfo{author}{P.~Bonnet}, \bibinfo{author}{A.~Joly},
\newblock \bibinfo{title}{Plant identification in an open-world (lifeclef 2016)},
\newblock in: \bibinfo{booktitle}{CLEF task overview 2016, CLEF: Conference and Labs of the Evaluation Forum, Sep. 2016, {\'E}vora, Portugal.}, \bibinfo{year}{2016}.
\bibitem[{Go\"{e}au et~al.(2017)Go\"{e}au, Bonnet, and Joly}]{plantclef2017}
\bibinfo{author}{H.~Go\"{e}au}, \bibinfo{author}{P.~Bonnet}, \bibinfo{author}{A.~Joly},
\newblock \bibinfo{title}{Plant identification based on noisy web data: the amazing performance of deep learning (lifeclef 2017)},
\newblock in: \bibinfo{booktitle}{CLEF task overview 2017, CLEF: Conference and Labs of the Evaluation Forum, Sep. 2017, Dublin, Ireland.}, \bibinfo{year}{2017}.
\bibitem[{Go{\"e}au et~al.(2018)Go{\"e}au, Bonnet, and Joly}]{expertlifeclef2018}
\bibinfo{author}{H.~Go{\"e}au}, \bibinfo{author}{P.~Bonnet}, \bibinfo{author}{A.~Joly},
\newblock \bibinfo{title}{Overview of expertlifeclef 2018: how far automated identification systems are from the best experts ?},
\newblock in: \bibinfo{booktitle}{CLEF task overview 2018, CLEF: Conference and Labs of the Evaluation Forum, Sep. 2018, Avignon, France.}, \bibinfo{year}{2018}.
\bibitem[{Go\"{e}au et~al.(2019)Go\"{e}au, Bonnet, and Joly}]{plantclef2019}
\bibinfo{author}{H.~Go\"{e}au}, \bibinfo{author}{P.~Bonnet}, \bibinfo{author}{A.~Joly},
\newblock \bibinfo{title}{Overview of lifeclef plant identification task 2019: diving into data deficient tropical countries},
\newblock in: \bibinfo{booktitle}{CLEF task overview 2019, CLEF: Conference and Labs of the Evaluation Forum, Sep. 2019, Lugano, Switzerland.}, \bibinfo{year}{2019}.
\bibitem[{Delprete and Feuillet(2013)}]{delprete2013marie}
\bibinfo{author}{P.~G. Delprete}, \bibinfo{author}{C.~Feuillet},
\newblock \bibinfo{title}{Marie-fran{\c{c}}oise pr{\'e}vost “fanchon”(1941--2013)},
\newblock \bibinfo{journal}{Taxon} \bibinfo{volume}{62} (\bibinfo{year}{2013}) \bibinfo{pages}{419--419}.
\bibitem[{Picek et~al.(2019)Picek, Sulc, and Matas}]{cmp2019lifeclef}
\bibinfo{author}{L.~Picek}, \bibinfo{author}{M.~Sulc}, \bibinfo{author}{J.~Matas},
\newblock \bibinfo{title}{Recognition of the amazonian flora by inception networks with test-time class prior estimation},
\newblock in: \bibinfo{booktitle}{CLEF working notes 2019, CLEF: Conference and Labs of the Evaluation Forum, Sep. 2019, Lugano, Switzerland.}, \bibinfo{year}{2019}.
\bibitem[{Chulif and Chang(2021)}]{NeuonAI2021}
\bibinfo{author}{S.~Chulif}, \bibinfo{author}{Y.~L. Chang},
\newblock \bibinfo{title}{Improved herbarium-field triplet network for cross-domain plant identification: Neuon submission to lifeclef 2021 plant},
\newblock in: \bibinfo{booktitle}{Working Notes of CLEF 2021 - Conference and Labs of the Evaluation Forum}, \bibinfo{year}{2021}.
\bibitem[{Youshan~Zhang(2021)}]{LehighPlantCLEF2021}
\bibinfo{author}{B.~D.~D. Youshan~Zhang},
\newblock \bibinfo{title}{Weighted pseudo labeling refinement for plant identification},
\newblock in: \bibinfo{booktitle}{Working Notes of CLEF 2021 - Conference and Labs of the Evaluation Forum}, \bibinfo{year}{2021}.
\bibitem[{Chulif and Chang(2020)}]{NeuonAI2020}
\bibinfo{author}{S.~Chulif}, \bibinfo{author}{Y.~L. Chang},
\newblock \bibinfo{title}{Herbarium-field triplets network for cross-domain plant identification - neuon submission to lifeclef 2020 plant},
\newblock in: \bibinfo{booktitle}{CLEF working notes 2020, CLEF: Conference and Labs of the Evaluation Forum, Sep. 2020, Thessaloniki, Greece.}, \bibinfo{year}{2020}.
\bibitem[{Sun and Saenko(2016)}]{sun2016deep}
\bibinfo{author}{B.~Sun}, \bibinfo{author}{K.~Saenko},
\newblock \bibinfo{title}{Deep coral: Correlation alignment for deep domain adaptation},
\newblock in: \bibinfo{booktitle}{European conference on computer vision}, \bibinfo{organization}{Springer}, \bibinfo{year}{2016}, pp. \bibinfo{pages}{443--450}.
\bibitem[{Villacis et~al.(2020)Villacis, Go\"{e}au, Bonnet, Mata-Montero, and Joly}]{ITCRPlantNet2020}
\bibinfo{author}{J.~Villacis}, \bibinfo{author}{H.~Go\"{e}au}, \bibinfo{author}{P.~Bonnet}, \bibinfo{author}{E.~Mata-Montero}, \bibinfo{author}{A.~Joly},
\newblock \bibinfo{title}{Domain adaptation in the context of herbarium collections: a submission to plantclef 2020},
\newblock in: \bibinfo{booktitle}{CLEF working notes 2020, CLEF: Conference and Labs of the Evaluation Forum, Sep. 2020, Thessaloniki, Greece.}, \bibinfo{year}{2020}.
\bibitem[{Motiian et~al.(2017)Motiian, Jones, Iranmanesh, and Doretto}]{fewshotda}
\bibinfo{author}{S.~Motiian}, \bibinfo{author}{Q.~Jones}, \bibinfo{author}{S.~Iranmanesh}, \bibinfo{author}{G.~Doretto},
\newblock \bibinfo{title}{Few-shot adversarial domain adaptation},
\newblock in: \bibinfo{booktitle}{Advances in Neural Information Processing Systems}, \bibinfo{year}{2017}, pp. \bibinfo{pages}{6670--6680}.

\end{thebibliography}
